\documentclass{article}

\usepackage{graphicx}
\graphicspath{ {./graphics/} }

\PassOptionsToPackage{square,numbers}{natbib}
\usepackage[final]{neurips_2022}

\usepackage{natbib}

\usepackage[utf8]{inputenc} % allow utf-8 input
\usepackage[T1]{fontenc}    % use 8-bit T1 fonts
\usepackage{hyperref}       % hyperlinks
\usepackage{url}            % simple URL typesetting
\usepackage{booktabs}       % professional-quality tables
\usepackage{amsfonts}       % blackboard math symbols
\usepackage{nicefrac}       % compact symbols for 1/2, etc.
\usepackage{microtype}      % microtypography
\usepackage{xcolor}         % colors
\usepackage[export]{adjustbox}
\usepackage{subcaption}
\usepackage{multirow}
\usepackage{paralist}

\title{System Safety Engineering for Social and Ethical ML Risks: A Case Study}
\author{
  Edgar W. Jatho III \\
  Naval Postgraduate School \\
  Monterey, CA \\
  \texttt{edgar.jatho@nps.edu } \\
  \And
   Logan O. Mailloux \\
   Naval Postgraduate School \\
   Monterey, CA \\
   \texttt{logan.mailloux@nps.edu} \\
  \And
   Shalaleh Rismani\\
   McGill University\\
   Montreal, Canada\\
   \texttt{shalaleh.rismani@mail.mcgill.ca} \\
  \And
   Eugene D. Williams \\
   Naval Postgraduate School \\
   Monterey, CA \\
   \texttt{eugene.williams@nps.edu} \\
  \And
   Joshua A. Kroll \\
   Naval Postgraduate School \\
   Monterey, USA \\
   \texttt{jkroll@nps.edu} \\
}

\begin{document}

\maketitle

\begin{abstract}
 Governments, industry, and academia have undertaken efforts to identify and mitigate harms in ML-driven systems, with a particular focus on social and ethical risks of ML components in complex sociotechnical systems. However, existing approaches are largely disjointed, ad-hoc and of unknown effectiveness. Systems safety engineering is a well established discipline with a track record of identifying and managing risks in many complex sociotechnical domains. We adopt the natural hypothesis that tools from this domain could serve to enhance risk analyses of ML in its context of use. To test this hypothesis, we apply a ``best of breed'' systems safety analysis, Systems Theoretic Process Analysis (STPA), to a specific high-consequence system with an important ML-driven component, namely the Prescription Drug Monitoring Programs (PDMPs) operated by many US States, several of which rely on an ML-derived risk score. We focus in particular on how this analysis can extend to identifying social and ethical risks and developing concrete design-level controls to mitigate them.
\end{abstract}

\section{Introduction}
A large and growing community of researchers, practitioners, and policymakers is concerned with the social and ethical risks that attend machine learning (ML) systems. These problems extend beyond the \emph{alignment} of the technology itself to the embodied and contextual use of ML-driven tools in entire sociotechnical systems. Systems safety engineering provides tools, techniques, and procedures that have been studied carefully in context for their ability to identify and control risks in complex sociotechnical systems~\cite{shin_stpa-based_2021, leveson_engineering_2016}. Although it has previously been suggested that such frameworks can perform similarly for ML systems safety risks, including social and ethical risks~\cite{dobbe_system_2022, raji_closing_2020}, %CITE: Others?
the concrete use of these techniques has not yet been validated. Nor is it known to what extent tools for managing risk in complex sociotechnical systems can be adapted to identifying social and ethical risks in particular. To test the hypothesis that systems safety engineering provides tools for assessing and mitigating social and ethical risk, we apply Leveson's System Theoretic Process Analysis (STPA) to a representative ML system, giving special attention to socially and ethically problematic system outcomes operationalized as safety hazards. As a case analysis, we apply STPA to a notional data-derived risk score as it would be used in the administration of the Prescription Drug Monitoring Program (PDMP) in many states. % Ed's footnote?
By analyzing the concrete application of STPA in a realistic ML case study, we can determine if STPA can effectively and repeatably identify and provide a path to eliminate or mitigate social and ethical risks that may result from a system.

 The state of the art in ML evaluation generally relies on ad-hoc review of chosen metrics such as AUC, metrics derived from confusion matrices, or -- for social and ethical risks -- so-called ``fairness metrics''. Metric-based evaluation is a fundamentally narrow view of model performance, especially for social and ethical risks: it frequently fails to address wider critical equities at stake~\cite{raji_ai_2021, malik_hierarchy_2020}. Fairness metrics, while a common proxy for identifying social and ethical concerns, are widely acknowledged to be imperfect operationalizations of underlying human values~\cite{mulligan_this_2019}. Additionally, it can be particularly difficult to assign responsibility for social and ethical risks in ML systems or to determine appropriate interventions to mitigate problems even once discovered~\cite{kroll_accountable_2017, raji_closing_2020, Cooper_2022, selbst_fairness_2019}.

In response, several efforts aim to create concrete evaluation frameworks designed to identify harms, especially social and ethical risks, and propose mitigation. For example, the US NIST's draft AI Risk Management Framework process~\cite{NIST_ai_2022} and the pending ``AI Act'' legislation in the European Union both categorize the management of social and ethical risks in ML systems as a risk management problem and envisage solutions in standardized evaluation frameworks. But even so systematized, assessments of social and ethical risks will remain ad-hoc -- these frameworks are based only on best consensus expert judgement. Instead, effective risk governance must be based in experience, scientific evaluation, and process validation. Practitioners and academics alike recognize the need for valid evaluation practices and welcome standardized frameworks~\cite{holstein_improving_2019, madaio_assessing_2022, rismani2023airplane, wilson_building_2021}.

%The field of system safety engineering has developed a number of
%tools validated by decades of experience and empirical research to
%discover risks and suggest effective controls for complex engineered
%systems~\cite{leveson_engineering_2016}. 
A strength of system safety engineering frameworks is that they connect abstract safety policies, which are difficult to make actionable through technical means alone, to implementable requirements. Tools from this domain further have the advantage of being regularly applied in high consequence domains, well studied, and providing a strong basis on which to systematize efforts to identify social and ethical risks~\cite{antoine_systems_2013, rising_systems-theoretic_2018}. Such tools include traditional safety-through-reliability techniques like fault-tree analysis (FTA)~\cite{lee_fault_1985} and Failure Mode and Effects Analysis (FMEA), quantification-oriented approaches used for decades to reduce the number of failures in systems under analysis~\cite{stamatis2003failure}. By contrast, Leveson's Systems Theoretic Accident Model and Process (STAMP)~\cite{leveson_engineering_2016} explicitly rejects the notion that reducing failures improves safety, noting that safety is a property of systems not components. STAMP instead models hazardous states that could lead to defined losses as insufficient control within an entire sociotechnical system.

Recognizing and responding to the social and ethical risks of an ML model requires viewing that tool in its context of use, as part of a broader sociotechnical system~\cite{martin_extending_2020}. We therefore borrow from STAMP its hazard analysis technique, Systems-Theoretic Process Analysis (STPA). STPA has a successful track record in high consequence domains~\cite{shin_stpa-based_2021, ishimatsu_hazard_2014}. By considering the full sociotechnical system, STPA contextualizes ML hazards with respect to social and ethical risks that result from component interactions and environmental factors in addition to component behaviors. Specifically, we apply STPA to a realistic notional case where social and ethical risks from ML have already been identified: several US state Prescription Drug Management Programs (PDMPs) use an ML-based risk score in their workflow. Our analysis seeks to answer several questions about STPA: Can STPA recover causal paths to social and ethical harms effectively? Does it suggest effective design interventions to avoid those unsafe system behaviors? What portions of the STPA process readily apply to social and ethical impact analysis and mitigation for ML systems? What aspects apply with only minor adjustments? What gaps still remain? Finally, we propose some adaptations or interpretations of STPA to bridge those identified gaps--towards a proven, systematic approach to advancing social and ethical impact analysis for complex ML-enabled sociotechnical systems. 

\section{Case Study: A Systems-Theoretic Process Analysis of PDMP Scoring}

STPA is a top-down system safety analysis tool, part of a family of techniques belonging to Leveson’s Systems Theoretic Accident Model and Process (STAMP) paradigm~\cite{leveson_engineering_2016, leveson_stpa_handbook_2018, shin_stpa-based_2021}. At a high level, STPA analysis requires four steps (for more detail about the STPA technique, see Appendix~\ref{STPA}):
\begin{inparaenum}[1)]
\item Defining the purpose of the analysis, including defining losses and hazards for the system of interest (SoI);
\item Modeling the full sociotechnical control structure for the system; and
\item Identifying unsafe control actions (UCAs) within each control loop which can cause a loss; and
\item Identifying causal loss scenarios for each UCA.
\end{inparaenum}

ML-based risk scores are widely used in Prescription Drug Monitoring Programs (PDMPs) throughout the United States.\footnote{The authors performed an informal search of state public health websites and official news releases and could confirm 27 of 51 states and D.C. use an ML based scoring system as a major component of their PDMP. Additionally, five of the top seven U.S. pharmacy businesses (by 2021 prescription revenue)~\cite{fein_pharmacy_2021}, likewise require their pharmacists to use an ML-based PDMP scoring system in their workflow. Many of these entities use the same scoring tool provided by a third-party vendor.} For more detailed information on our subject ML system, refer to Appendix \ref{PDMP}. We examine this system using STPA to better understand how social and ethical harms arise from the ML components and to identify design constraints and corrective controls to mitigate them.

In step one of the STPA, the team defined the purpose of our analysis as to identify and where possible, eliminate sociotechnical harms whether manifested as various forms of inequity between social groups, such as representational, allocative, quality of service, interpersonal or societal harms~\cite{shelby2023harms}, or more narrowly manifested as traditional safety harms such as loss of life, injury, and loss of quality of life to individuals. For more detail on the harms taxonomy we adopt for our analysis, see Appendix \ref{harms}. This part of the process serves to align the resulting system with desired societal values~\cite{hendrycks_unsolved_2022}. We identified our primary stakeholders as patients, followed by doctors and pharmacists. %, noting everyone experiences the role of a patient. 
Our initial list of losses was built from the harms stated above and included approximately 30 losses (Appendix \ref{initial_loss}), which we consolidated and further grouped into a set of five distilled losses (Appendix \ref{distilled} for the grouping). We found the PDMP risk score losses were best grouped thus: 1) Death, Injury or Disability, 2) Disparity of Benefits or Harms, 3) Social or Economic Injury, 4) Damage to Quality of Healthcare, 5) Coerced Criminality or Unsafe Treatment.

%\textbf{PDMP System Losses}
%\begin{enumerate}
%    \item Death, Injury or Disability
%    \item Disparity of Benefits/Harms
%    \item Social/Economic Injury
%    \item Damage to Quality of Healthcare
%    \item Forced Criminality or Unsafe Treatment
%\end{enumerate}
%It was during this phase, that we first engaged with the idea of the PDMP score's overall objective.  As it is considered a clinical tool by physicians and pharmacists, our intuition is that its first priority should be the preservation of life, restoration of health and to "do no harm." As we explored the various state manuals, the intended users, the data features specified as critical (i.e., number of pharmacies visited, number of prescribers in the last two years, method of payment, maximum morphine milligram equivalency (MME) in the last year, diagnoses, etc)~\cite{LA_pharm, texas_narx_manual}, the future data sets considered promising to provide improvements (criminal justice)
From these losses a set of 10 system-specific hazards were identified, see table \ref{haz_tbl} in Appendix \ref{hazards}. For step two, the team continued to comb through documentation and research on PDMP risk score ML models, seeking to develop a useful control structure for analysis. Early on, we acknowledged a need to address both the operational control structure \textit{context} of the ML system as well as solve how and to what level of abstraction the ML life-cycle should be modeled within the control structure.  This was an iterative process and settled on the structures found in Appendix~\ref{figs}, figure~\ref{ovrall_W_inset} for the operational context and figure~\ref{cycle-treatment} for the treatment of the ML life-cycle. The usefulness of this interpretation will be discussed further in section~\ref{DD_expl}. Step two also required defining each controller and controlled process' function, the associated control actions and feedback as depicted in figures \ref{care} and \ref{data}.\footnote{Referenced descriptions within the control structure are treated in more detail in Appendix \ref{cont_det}} 

For STPA step three, the team used the control structure and potential hazards to identify control actions whose misapplication can lead to a hazard. These are unsafe control actions (UCAs), and for each control loop, we considered the four standard misapplications of control:
\begin{inparaenum}[1)]
\item Not providing control causes hazard;
\item providing control causes hazard; 
\item control is too early, too late, or out of order;
\item incorrect control duration (too long, too short).
\end{inparaenum}
This yielded the control action tables found in Appendix \ref{UCA}. We found that when applying the four standard control misapplications to the ML development cycle control loop, whereas the temporal question (i.e., too early, too late, wrong order) was not useful, it was useful to alter the duration misapplication (too long, too short) to instead refer to concepts of quantity, i.e., too many/much, too few/little. 
Finally, with the UCAs outlined, the causal scenarios logically follow, manifesting system losses.  The result is that logical requirements for sociotechnical system design modifications to eliminate or mitigate the risks of the identified hazards become far easier to motivate, reason about and outline.

\subsection{PDMP Findings}
This analysis resulted in an informative contextualized control structure for a PDMP risk score algorithm and its surrounding sociotechnical system. It identified 30 specific initial losses (Appendix \ref{initial_loss}) which were grouped and pared down to five loss types (Appendix \ref{distilled}). The analysis was scoped to the two most interesting of the seven primary control loops identified and modeled, namely \textit{Patient Care}, figure \ref{care} and \textit{Data Decisions}, figure \ref{data}. Analysis provided a combined 13 unsafe control actions as shown in tables \ref{UCA_care} and \ref{UCA_dev} to inform and motivate actionable sociotechnical system requirements to eliminate or mitigate the identified harms in the system going forward.

Many calls for action to adopt PDMP risk scores were motivated by the U.S. opioid epidemic, the tragic and increasing loss of life fueled by opioid overdoses every year. One early discovery that quickly became apparent in the STPA analysis was that, though often motivated by the desire to save lives and prevent opioid overdoses and Opioid Use Disorder (OUD), the objective function of this ML model is aimed at minimizing drug \textit{diversion} instead, defined by the US Department of Health and Human Services as "the illegal distribution or abuse of prescription drugs or their use for purposes not intended by the prescriber." The underlying premise is that reducing drug diversion will save lives.  However, considering the hazards, control loops, and UCAs we identified in our short analysis of this system, it is rationally defensible to hold that the opposite could be true\footnote{A possibility at least not refuted by nationwide statistics showing a marked reduction in drug diversion and opioid prescribing coinciding with the adoption of PDMPs in every state, mandated risk score use in over half, and an overall overdose death rate that is nevertheless increasing~\cite{NIH_overdose_2022}}. The control loops show possible mechanisms whereby actions taken by prescribers and dispensers to ostensibly protect an individual, or act in their business's best interest based on the patient's score, instead may result in more desperate and dangerous behavior by the patient as effective treatment is sought but may not be provided. This overall issue highlights a frequent problem in ML systems where the objective function may not be suitably matched to the system goal~\cite{obermeyer_dissecting_2019}, here immediately raised as a possible issue in the first steps of the analysis and further questioned as the STPA is carried out.

% **** Below Paragraph is removed because it needs to be rewritten:
% Goal of rewriting: Point out that a relevant consideration of identified hazardous states is how often the hazardous states are entered. 
% Additionally, our STPA highlighted the losses introduced by the PDMP's resulting false positive and false negative rates. As part of ongoing work, Kilby found that systems based on the same datasets and features used in the most ubiquitous PDMP risk score systems, were shown to have very high false positive and false negative rates. This remained true even when threshold scores were held to be the top 1 percent of scores assigned patients. Stated differently, of those identified by the ML system to be at the top 1\% highest risk only 11\% were true positives, 89\% in this category were false positives.  Additionally, this system resulted in only 37\% of the total true positive patients (those needing help) being identified by the algorithm for help~\cite{Kilby_algorithm, kilby_economics_2016}. This leaves open the important question of acceptable thresholds of false positive and false negative rates for this system. Additionally, what controls are necessary to deal with the consequences and resulting harms of these high error rates?

\subsection{ML Life Cycle: Data Decision Phase Example}\label{DD_expl}
One challenge of applying STPA to machine learning that was identified early was how to best apply control structure and follow-on analysis to the development life-cycle of ML systems. The team made progress on this challenge, coming to the conclusion that the various phases of the life-cycle should be abstracted as individual control loops with the \textit{phase} modeled as the \textit{controlled process} and the \textit{development team} modeled as the \textit{controller} (figs \ref{cycle-treatment} and \ref{data}). Our effort shows this to be a promising way forward having resulted in control actions, meaningful validation questions and requirements with respect to the social and ethical impact framing. With respect to PDMP risk scores, table \ref{UCA_dev} shows our results in finding unsafe control actions in the data decision phase alone that can lead to social and ethical losses.  This is an ongoing investigation with new insights, developments and applications forthcoming.

\section{Discussion}
Conducting an STPA for social and ethical impact forces development teams and other stakeholders to do the necessary work to fully consider the larger system context that a product or component will inhabit, the larger sociotechnical system.  As stated before, recent history shows that frequently, teams are hyper focused on the SoI they are developing and thus miss the bigger picture, the larger purpose and direction. The STPA process cultivates a rich appreciation for the sociotechnical system a new system or product is entering, and provides an effective abstraction with which to reason about these harms, as well as the instruments, tools, and methods (both social and technical) that we can bring to bear to eliminate or mitigate them.

Another important contribution of STPA is the mandate to consider carefully the overall goal of the system (step 1) and in our treatment, verify that the objective function that is adopted does not itself lead to the social and ethical losses the STPA identifies. For example, applying STPA to the health benefits scoring system system studied by Obermeyer~\cite{obermeyer_dissecting_2019} could reveal the need for a check for racial and socio-economic disparities resulting from an objective function mismatch early in development and could have enabled a shift to a more appropriate governing optimization. 

Finding a way to abstract and model STPA in the ML life-cycle was a challenging aspect of applying STPA to S\&E impact in ML systems.  One benefit of the ML life-cycle treatment the team developed (see section~\ref{DD_expl} and figures \ref{cycle-treatment} and~\ref{data}) is that it breaks the STPA modeling into manageable pieces where each phase or stage has a reasonable number of control actions which can be decomposed into measurable and verifiable considerations and checks.

A particular nuance of a thorough social and ethical impact STPA should also consider whether the resulting product algorithm could be used by the funding or owning company (or sold as a service to another) in a manner not overtly intended in the stated purpose of the system but (regardless of legality) presents a potential negative social or ethical impact on society. Part of these analyses should also be to consider these types of uses or outputs to other interests and suggest mitigations or scope statements that warn of these potential issues and the need for separate STPA analysis for those uses. 

In the case study system, a number of the controllers or controlled processes involved humans who are free agents, such as doctors, pharmacists and their staffs.  Considering Goodhart's law "...when a measure becomes a target, it ceases to be a good measure," it is to be expected that an entity modeled as a controlled process may alter their behavior or feedback paths to exert influence on the modeled controller.  One potential approach would be to also examine these relationships in the opposite direction and ask the same control, hazards, and scenario questions. This practice can reveal potential for, in the case of PDMP score systems, such behaviors as abandonment by doctors and service refusal by pharmacists, and the attendant losses which may result. Identification of these types of potential behaviors only serve to expand the analytical potential for discovering hazards and thus identifying controls to prevent those hazards.

One final benefit of the STPA process  is that it necessarily provides a traceable path from every derived requirement to its causal scenarios, contributing unsafe control actions, control loops of origin and orginating hazards and losses. This property enables the complete retracing of the logic and reasoning behind every decision in design and operations.

\subsection{Conclusion}
In this paper we record the results of a two week sprint where we applied STPA, a traditional system safety engineering analysis methodology to the challenge of assessing the social and ethical impact of a machine learning system, a Prescription Drug Monitoring Program risk score. We found that STPA's rigorous approach when coupled with a thorough harms taxonomy produced a trove of hazards and unsafe control actions against which new system requirements for sociotechnical control mechanisms could subsequently be applied to prevent social and ethical losses. Additionally, we adapted a useful abstraction of the machine learning life cycle for STPA which recovered potential unsafe control actions in a manner similar to those captured by their analogous operational control loops.  Future work will examine the applicability and generalizability of STPA for social and ethical impact by investigating case studies across different applications of ML systems.  In so doing we seek to introduce a tool to provide an organized, proven systematic approach to social and ethical analysis for complex ML-enabled sociotechnical systems.

\subsection*{Acknowledgements}
This work would not have been possible without detailed input from and discussions with Renee Shelby, Negar Rostamzadeh, and Andrew Smart of Google. We are also grateful to Abigail Z. Jacobs for encouraging the pursuit of systems safety tools for assessing ML risks.

\medskip

\bibliography{My_Library}

\begin{thebibliography}{43}
\providecommand{\natexlab}[1]{#1}
\providecommand{\url}[1]{\texttt{#1}}
\expandafter\ifx\csname urlstyle\endcsname\relax
  \providecommand{\doi}[1]{doi: #1}\else
  \providecommand{\doi}{doi: \begingroup \urlstyle{rm}\Url}\fi

\bibitem[Shin et~al.(2021)Shin, Lee, Shin, Jang, and
  Park]{shin_stpa-based_2021}
Sung-Min Shin, Sang~Hun Lee, Seung K~I Shin, Inseok Jang, and Jinkyun Park.
\newblock {STPA}-{Based} {Hazard} and {Importance} {Analysis} on {NPP} {Safety}
  {I}\&{C} {Systems} {Focusing} on {Human}–{System} {Interactions}.
\newblock \emph{Reliab. Eng. Syst. Saf.}, 213:\penalty0 107698, September 2021.

\bibitem[Leveson(2016)]{leveson_engineering_2016}
Nancy~G Leveson.
\newblock \emph{Engineering a safer world: {Systems} thinking applied to
  safety}.
\newblock The MIT Press, 2016.

\bibitem[Dobbe(2022)]{dobbe_system_2022}
Roel I.~J. Dobbe.
\newblock System {Safety} and {Artificial} {Intelligence}, February 2022.
\newblock URL \url{http://arxiv.org/abs/2202.09292}.
\newblock arXiv:2202.09292 [cs, eess].

\bibitem[Raji et~al.(2020)Raji, Smart, White, Mitchell, Gebru, Hutchinson,
  Smith-Loud, Theron, and Barnes]{raji_closing_2020}
Inioluwa~Deborah Raji, Andrew Smart, Rebecca~N White, Margaret Mitchell, Timnit
  Gebru, Ben Hutchinson, Jamila Smith-Loud, Daniel Theron, and Parker Barnes.
\newblock Closing the {AI} {Accountability} {Gap}: {Defining} an {End}-to-{End}
  {Framework} for {Internal} {Algorithmic} {Auditing}.
\newblock \emph{ACM Conference on Fairness, Accountability, and Transparency},
  2020.

\bibitem[Raji et~al.(2021)Raji, Bender, Paullada, Denton, and
  Hanna]{raji_ai_2021}
Inioluwa~Deborah Raji, Emily~M Bender, Amandalynne Paullada, Emily Denton, and
  Alex Hanna.
\newblock {AI} and the everything in the whole wide world benchmark.
\newblock \emph{arXiv preprint arXiv:2111.15366}, 2021.

\bibitem[Malik(2020)]{malik_hierarchy_2020}
Momin~M. Malik.
\newblock A {Hierarchy} of {Limitations} in {Machine} {Learning}, 2020.
\newblock \_eprint: 2002.05193.

\bibitem[Mulligan et~al.(2019)Mulligan, Kroll, Kohli, and
  Wong]{mulligan_this_2019}
Deirdre~K. Mulligan, Joshua~A. Kroll, Nitin Kohli, and Richmond~Y. Wong.
\newblock This {Thing} {Called} {Fairness}: {Disciplinary} {Confusion}
  {Realizing} a {Value} in {Technology}.
\newblock \emph{Proceedings of the ACM on Human-Computer Interaction},
  3\penalty0 (CSCW):\penalty0 119:1--119:36, November 2019.
\newblock \doi{10.1145/3359221}.
\newblock URL \url{https://doi.org/10.1145/3359221}.

\bibitem[Kroll et~al.(2017)Kroll, Huey, Barocas, Felten, Reidenberg, Robinson,
  and Yu]{kroll_accountable_2017}
Joshua~A. Kroll, Joanna Huey, Solon Barocas, Edward~W. Felten, Joel~R.
  Reidenberg, David~G. Robinson, and Harlan Yu.
\newblock Accountable {Algorithms}.
\newblock \emph{University of Pennsylvania Law Review}, 165\penalty0 (3), 2017.

\bibitem[Cooper et~al.(2022)Cooper, Moss, Laufer, and Nissenbaum]{Cooper_2022}
A.~Feder Cooper, Emanuel Moss, Benjamin Laufer, and Helen Nissenbaum.
\newblock Accountability in an algorithmic society: Relationality,
  responsibility, and robustness in machine learning.
\newblock In \emph{2022 {ACM} Conference on Fairness, Accountability, and
  Transparency}. {ACM}, jun 2022.
\newblock \doi{10.1145/3531146.3533150}.
\newblock URL \url{https://doi.org/10.1145%2F3531146.3533150}.

\bibitem[Selbst et~al.(2019)Selbst, Boyd, Friedler, Venkatasubramanian, and
  Vertesi]{selbst_fairness_2019}
Andrew~D Selbst, Danah Boyd, Sorelle~A Friedler, Suresh Venkatasubramanian, and
  Janet Vertesi.
\newblock Fairness and abstraction in sociotechnical systems.
\newblock In \emph{Proceedings of the conference on fairness, accountability,
  and transparency}, pages 59--68, 2019.

\bibitem[NIS(2022)]{NIST_ai_2022}
{AI} {Risk} {Management} {Framework} {\textbar} {NIST}, August 2022.
\newblock URL
  \url{https://www.nist.gov/system/files/documents/2022/08/18/AI_RMF_2nd_draft.pdf}.

\bibitem[Holstein et~al.(2019)Holstein, Wortman~Vaughan, Daumé, Dudik, and
  Wallach]{holstein_improving_2019}
Kenneth Holstein, Jennifer Wortman~Vaughan, Hal Daumé, Miro Dudik, and Hanna
  Wallach.
\newblock Improving {Fairness} in {Machine} {Learning} {Systems}: {What} {Do}
  {Industry} {Practitioners} {Need}?
\newblock In \emph{Proceedings of the 2019 {CHI} {Conference} on {Human}
  {Factors} in {Computing} {Systems}}, {CHI} '19, pages 1--16, New York, NY,
  USA, 2019. Association for Computing Machinery.
\newblock ISBN 978-1-4503-5970-2.
\newblock \doi{10.1145/3290605.3300830}.
\newblock URL \url{https://doi.org/10.1145/3290605.3300830}.
\newblock event-place: Glasgow, Scotland Uk.

\bibitem[Madaio et~al.(2022)Madaio, Egede, Subramonyam, Vaughan, and
  Wallach]{madaio_assessing_2022}
Michael Madaio, Lisa Egede, Hariharan Subramonyam, Jennifer~Wortman Vaughan,
  and Hanna Wallach.
\newblock Assessing the {Fairness} of {AI} {Systems}: {AI} {Practitioners}'
  {Processes}, {Challenges}, and {Needs} for {Support}, 2022.
\newblock Issue: CSCW1 Pages: 1–26 Publication Title: Proceedings of the ACM
  on Human-Computer Interaction Volume: 6.

\bibitem[Rismani et~al.(2022)Rismani, Shelby, Smart, Jatho, Kroll, Moon, and
  Rostamzadeh]{rismani2023airplane}
Shalaleh Rismani, Renee Shelby, Andrew Smart, Edgar Jatho, Joshua Kroll, AJung
  Moon, and Negar Rostamzadeh.
\newblock From plane crashes to algorithmic harm: applicability of safety
  engineering frameworks for responsible ml.
\newblock 2022.
\newblock Submitted to the ACM Conference on Human Factors in Computing Systems
  (CHI) 2023, awaiting response.

\bibitem[Wilson et~al.(2021)Wilson, Ghosh, Jiang, Mislove, Baker, Szary,
  Trindel, and Polli]{wilson_building_2021}
Christo Wilson, Avijit Ghosh, Shan Jiang, Alan Mislove, Lewis Baker, Janelle
  Szary, Kelly Trindel, and Frida Polli.
\newblock Building and {Auditing} {Fair} {Algorithms}: {A} {Case} {Study} in
  {Candidate} {Screening}.
\newblock In \emph{Proceedings of the 2021 {ACM} {Conference} on {Fairness},
  {Accountability}, and {Transparency}}, {FAccT} '21, pages 666--677, New York,
  NY, USA, 2021. Association for Computing Machinery.
\newblock ISBN 978-1-4503-8309-7.
\newblock \doi{10.1145/3442188.3445928}.
\newblock URL \url{https://doi.org/10.1145/3442188.3445928}.
\newblock event-place: Virtual Event, Canada.

\bibitem[Antoine(2013)]{antoine_systems_2013}
Blandine Antoine.
\newblock \emph{Systems {Theoretic} {Hazard} {Analysis} ({STPA}) applied to the
  risk review of complex systems : an example from the medical device
  industry}.
\newblock Thesis, Massachusetts Institute of Technology, 2013.
\newblock URL \url{https://dspace.mit.edu/handle/1721.1/79424}.
\newblock Accepted: 2013-07-09T19:30:13Z.

\bibitem[Rising and Leveson(2018)]{rising_systems-theoretic_2018}
John~M. Rising and Nancy~G. Leveson.
\newblock Systems-{Theoretic} {Process} {Analysis} of space launch vehicles.
\newblock \emph{Journal of Space Safety Engineering}, 5\penalty0 (3):\penalty0
  153--183, September 2018.
\newblock ISSN 2468-8967.
\newblock \doi{10.1016/j.jsse.2018.06.004}.
\newblock URL
  \url{https://www.sciencedirect.com/science/article/pii/S2468896718300296}.

\bibitem[Lee et~al.(1985)Lee, Grosh, Tillman, and Lie]{lee_fault_1985}
W.~S. Lee, D.~L. Grosh, F.~A. Tillman, and C.~H. Lie.
\newblock Fault {Tree} {Analysis}, {Methods}, and {Applications} {A} {Review}.
\newblock \emph{IEEE Transactions on Reliability}, R-34\penalty0 (3):\penalty0
  194--203, August 1985.
\newblock ISSN 1558-1721.
\newblock \doi{10.1109/TR.1985.5222114}.
\newblock Conference Name: IEEE Transactions on Reliability.

\bibitem[Stamatis(2003)]{stamatis2003failure}
Diomidis~H Stamatis.
\newblock \emph{Failure mode and effect analysis: FMEA from theory to
  execution}.
\newblock Quality Press, 2003.

\bibitem[Jr et~al.(2020)Jr, Prabhakaran, Kuhlberg, Smart, and
  Isaac]{martin_extending_2020}
Donald~Martin Jr, Vinodkumar Prabhakaran, Jill Kuhlberg, Andrew Smart, and
  William~S. Isaac.
\newblock Extending the {Machine} {Learning} {Abstraction} {Boundary}: {A}
  {Complex} {Systems} {Approach} to {Incorporate} {Societal} {Context}, 2020.
\newblock \_eprint: 2006.09663.

\bibitem[Ishimatsu et~al.(2014)Ishimatsu, Leveson, Thomas, Fleming, Katahira,
  Miyamoto, Ujiie, Nakao, and Hoshino]{ishimatsu_hazard_2014}
Takuto Ishimatsu, Nancy~G. Leveson, John~P. Thomas, Cody~H. Fleming, Masafumi
  Katahira, Yuko Miyamoto, Ryo Ujiie, Haruka Nakao, and Nobuyuki Hoshino.
\newblock Hazard {Analysis} of {Complex} {Spacecraft} {Using}
  {Systems}-{Theoretic} {Process} {Analysis}.
\newblock \emph{MIT web domain}, February 2014.
\newblock ISSN 0022-4650.
\newblock URL \url{https://dspace.mit.edu/handle/1721.1/96964}.
\newblock Accepted: 2015-05-12T16:51:38Z Publisher: American Institute of
  Aeronautics and Astronautics.

\bibitem[Leveson and Thomas(2018)]{leveson_stpa_handbook_2018}
Nancy Leveson and John Thomas.
\newblock {STPA}\_handbook, March 2018.

\bibitem[Fein and {Ph.D.}(2021)]{fein_pharmacy_2021}
Adam~J. Fein and {Ph.D.}
\newblock The {Top} 15 {U}.{S}. {Pharmacies} of 2021: {Market} {Shares} and
  {Revenues} at the {Biggest} {Companies}, 2021.
\newblock URL
  \url{https://www.drugchannels.net/2022/03/the-top-15-us-pharmacies-of-2021-market.html}.

\bibitem[Shelby et~al.(2022)Shelby, Rismani, Henee, Moon, Rostamzadeh,
  Nicholas, Yilla, Gallegos, Smart, Garcia, and Virk]{shelby2023harms}
Renee Shelby, Shalaleh Rismani, Kathryn Henee, AJung Moon, Negar Rostamzadeh,
  Paul Nicholas, N'Mah Yilla, Jess Gallegos, Andrew Smart, Emilio Garcia, and
  Gurleen Virk.
\newblock Sociotechnical harms: Scoping a taxonomy for harm reduction.
\newblock 2022.
\newblock Submitted to the ACM Conference on Human Factors in Computing Systems
  (CHI) 2023, awaiting response.

\bibitem[Hendrycks et~al.(2022)Hendrycks, Carlini, Schulman, and
  Steinhardt]{hendrycks_unsolved_2022}
Dan Hendrycks, Nicholas Carlini, John Schulman, and Jacob Steinhardt.
\newblock Unsolved {Problems} in {ML} {Safety}, June 2022.
\newblock URL \url{http://arxiv.org/abs/2109.13916}.
\newblock arXiv:2109.13916 [cs].

\bibitem[Abuse(2022)]{NIH_overdose_2022}
National Institute on~Drug Abuse.
\newblock Overdose {Death} {Rates}, January 2022.
\newblock URL
  \url{https://nida.nih.gov/research-topics/trends-statistics/overdose-death-rates}.

\bibitem[Obermeyer et~al.(2019)Obermeyer, Powers, Vogeli, and
  Mullainathan]{obermeyer_dissecting_2019}
Ziad Obermeyer, Brian Powers, Christine Vogeli, and Sendhil Mullainathan.
\newblock Dissecting racial bias in an algorithm used to manage the health of
  populations.
\newblock \emph{Science}, 366\penalty0 (6464):\penalty0 447--453, 2019.
\newblock ISSN 0036-8075.
\newblock \doi{10.1126/science.aax2342}.
\newblock URL \url{https://science.sciencemag.org/content/366/6464/447}.
\newblock Publisher: American Association for the Advancement of Science
  \_eprint: https://science.sciencemag.org/content/366/6464/447.full.pdf.

\bibitem[Perrow(1984)]{perrow_normal_1984}
Charles Perrow.
\newblock \emph{Normal accidents: {Living} with high risk technologies}.
\newblock Basic Books, New York, 1984.

\bibitem[Thomas(2021)]{thomas_2021}
John Thomas.
\newblock Introduction to stpa, 2021.
\newblock URL \url{https://www.youtube.com/watch?v=2W-iqnPbhyc}.
\newblock Engineering Systems Lab, MIT.

\bibitem[Kilby(2021)]{Kilby_algorithm}
Angela~E. Kilby.
\newblock Algorithmic fairness in predicting opioid use disorder using machine
  learning.
\newblock In \emph{Proceedings of the 2021 ACM Conference on Fairness,
  Accountability, and Transparency}, FAccT '21, page 272, New York, NY, USA,
  2021. Association for Computing Machinery.
\newblock ISBN 9781450383097.
\newblock \doi{10.1145/3442188.3445891}.
\newblock URL \url{https://doi.org/10.1145/3442188.3445891}.

\bibitem[Ind(2020)]{indiana_narx_manual}
\emph{Indiana Prescription Monitoring Program: Requestor User Support Manual}.
\newblock Indiana Board of Pharmacy Prescription Monitoring Program (INSPECT),
  9901 Linn Station Road Louisville, KY 40223, version 2.1 edition, 2020.
\newblock URL
  \url{https://www.in.gov/pla/inspect/files/Narxcare_user_guide.pdf}.

\bibitem[App(2021)]{texas_narx_manual}
\emph{Texas Prescription Monitoring Program: Requestor User Support Manual}.
\newblock Appriss Health, 9901 Linn Station Road Louisville, KY 40223, version
  2.5 edition, 2021.

\bibitem[Health(2018{\natexlab{a}})]{NC_narx}
Appriss Health.
\newblock N.{C}. {Department} of {Health} and {Human} {Services} {Partners}
  with {Appriss} {Health} to {Provide} {Access} to {NarxCare} {Platform} and
  {Controlled} {Substances} {Reporting} {System} {Information} {Directly}
  within {Electronic} {Health} {Record} and {Pharmacy} {Management} {Systems},
  November 2018{\natexlab{a}}.
\newblock URL \url{https://tinyurl.com/accesstoNarxCare2}.

\bibitem[nar()]{narxcare_webpage}
{NarxCare} {\textbar} {Substance} {Abuse} {Software} {Solution}.
\newblock URL \url{https://bamboohealth.com/solutions/narxcare/}.

\bibitem[Health(2018{\natexlab{b}})]{west_virginia_2018}
Appriss Health.
\newblock West {Virginia} {Partners} with {Appriss} {Health} to {Enable}
  {Access} to {NarxCare} {Substance} {Use} {Disorder} {Platform} {Directly}
  within {Clinical} {Workflow} for {All} {Prescribers} and {Pharmacists}
  {State}-{Wide}, June 2018{\natexlab{b}}.
\newblock URL \url{https://tinyurl.com/accesstoNarxCare}.

\bibitem[Kilby(2016)]{kilby_economics_2016}
Angela~E. Kilby.
\newblock \emph{The economics of pain management}.
\newblock {PhD} {Thesis}, Department of Economics, Massachusetts Institute of
  Technology., Cambridge, MA, USA, 2016.
\newblock URL \url{http://hdl.handle.net/1721.1/107321}.

\bibitem[Bender et~al.(2021)Bender, Gebru, McMillan-Major, and
  Shmitchell]{bender_dangers_2021}
Emily~M. Bender, Timnit Gebru, Angelina McMillan-Major, and Shmargaret
  Shmitchell.
\newblock On the {Dangers} of {Stochastic} {Parrots}: {Can} {Language} {Models}
  {Be} {Too} {Big}?
\newblock In \emph{Proceedings of the 2021 {ACM} {Conference} on {Fairness},
  {Accountability}, and {Transparency}}, {FAccT} '21, pages 610--623, New York,
  NY, USA, 2021. Association for Computing Machinery.
\newblock ISBN 978-1-4503-8309-7.
\newblock \doi{10.1145/3442188.3445922}.
\newblock URL \url{https://doi.org/10.1145/3442188.3445922}.
\newblock event-place: Virtual Event, Canada.

\bibitem[Kroll(2020)]{kroll_accountability_2020}
Joshua~A. Kroll.
\newblock Accountability in {Computer} {Systems}.
\newblock In Markus~D. Dubber, Frank Pasquale, and Sunit Das, editors,
  \emph{The {Oxford} {Handbook} of the {Ethics} of {AI}}. Oxford University
  Press, Oxford, 2020.
\newblock URL
  \url{https://papers.ssrn.com/sol3/papers.cfm?abstract_id=3608468}.
\newblock Section: 9.

\bibitem[Martin et~al.(2020)Martin, Prabhakaran, Kuhlberg, Smart, and
  Isaac]{martin_participatory_2020}
Donald Martin, Jr, Vinodkumar Prabhakaran, Jill Kuhlberg, Andrew Smart, and
  William~S Isaac.
\newblock Participatory {Problem} {Formulation} for {Fairer} {Machine}
  {Learning} {Through} {Community} {Based} {System} {Dynamics}.
\newblock May 2020.
\newblock \_eprint: 2005.07572.

\bibitem[Heikklä(2022)]{politico_dutch_2022}
Melissa Heikklä.
\newblock Dutch scandal serves as a warning for {Europe} over risks of using
  algorithms, March 2022.
\newblock URL \url{https://tinyurl.com/dutchscandal}.

\bibitem[Perkowitz(2021)]{perkowitz_bias_2021}
Sidney Perkowitz.
\newblock The {Bias} in the {Machine}: {Facial} {Recognition} {Technology} and
  {Racial} {Disparities}.
\newblock \emph{MIT Case Studies in Social and Ethical Responsibilities of
  Computing}, \penalty0 (Winter 2021), February 2021.
\newblock \doi{10.21428/2c646de5.62272586}.
\newblock URL \url{https://mit-serc.pubpub.org/pub/bias-in-machine/release/1}.
\newblock Publisher: MIT Schwarzman College of Computing.

\bibitem[Julia~Angwin and Kirchner(2016)]{Angwin_arrest_recidivism_2016}
Surya~Mattu Julia~Angwin, Jeff~Larson and Lauren Kirchner.
\newblock Machine {Bias} — {ProPublica}, May 2016.
\newblock URL
  \url{https://www.propublica.org/article/machine-bias-risk-assessments-in-criminal-sentencing}.

\bibitem[Li and Chignell(2022)]{li_fmea-ai_2022}
Jamy Li and Mark Chignell.
\newblock {FMEA}-{AI}: {AI} fairness impact assessment using failure mode and
  effects analysis.
\newblock \emph{AI and Ethics}, March 2022.
\newblock Publisher: Springer Nature.

\end{thebibliography}

\appendix
\section{System Theoretic Process Analysis (STPA)}\label{STPA}

STPA is a top-down system safety analysis tool, part of a family of techniques belonging to Leveson’s Systems Theoretic Accident Model and Process (STAMP) paradigm~\cite{leveson_engineering_2016}. STPA is a proven and systematic hazard analysis process that frames accidents not in terms of component failures but instead as a control structure which prevents the subject system from entering hazardous states which could lead to unacceptable losses. Typically, losses are defined by key stakeholders such as system owners or operators and often include: death, injury, damage to property, financial loss, or loss of mission. Hazards are typically described as a system state that when coupled with a specific set of worst-case conditions results in a loss(es). In STPA, hazards and their resulting losses follow from inadequate system control, i.e., unsafe control actions (UCA)s.~\cite{leveson_engineering_2016}. 

Notably, STPA expands on the scope of traditional reliability-driven safety analysis to include unforeseen behavioral interactions across the entire sociotechnical system. For example, STPA analysis includes consideration of human-machine interfaces, supporting governance hierarchies, and even organizational culture. Rejecting the idea of a “root-cause”, STPA proceeds from the thinking that accidents are, as Perrow puts it, “normal behaviors” of complex systems. Far from unlikely, accidents are inevitable emergent system behaviors, arising from system structure and function that must be identified and subsequently controlled \cite{perrow_normal_1984, leveson_engineering_2016}. 

\subsection{STPA Process}
Applying STPA consists of the following core steps which are intended to be applied while studying the system of interest across the system’s lifecycle. The STPA process should be repeated at higher levels of detail until the purpose of the analysis can be satisfactorily addressed. 
\begin{enumerate} 
    \item Define the purpose of the analysis: Identify stakeholders, define what constitutes a loss and surface system-specific hazards to be eliminated or mitigated to prevent losses from occuring. 
    \item Model the full sociotechnical control structure for the system. This involves mapping the feedback control loops of the sociotechnical system to the level of abstraction necessary to meaningfully reason about them.
    \item Considering the control structure and potential hazards, identify unsafe control actions (UCAs) for each control loop, (i.e., what controller action, inaction, or misapplied action (too early, too late), or applied for the wrong duration, etc. -  can go wrong and cause a loss?)
    \item Identify and consider potential loss scenarios (i.e. causal scenarios) for each UCA. A tangible benefit of this final step is the development of a set of requirements that need to be enforced to ensure a safe sociotechnical system results, these may include but are not limited to new design decisions, requirements, procedures, operator training, test cases or even periodic audits~\cite{leveson_stpa_handbook_2018}.
\end{enumerate}

\section{Control Structure}\label{figs}
\begin{figure}[h!]

\begin{subfigure}{0.5\textwidth}
\includegraphics[width=0.9\linewidth, height=6cm]{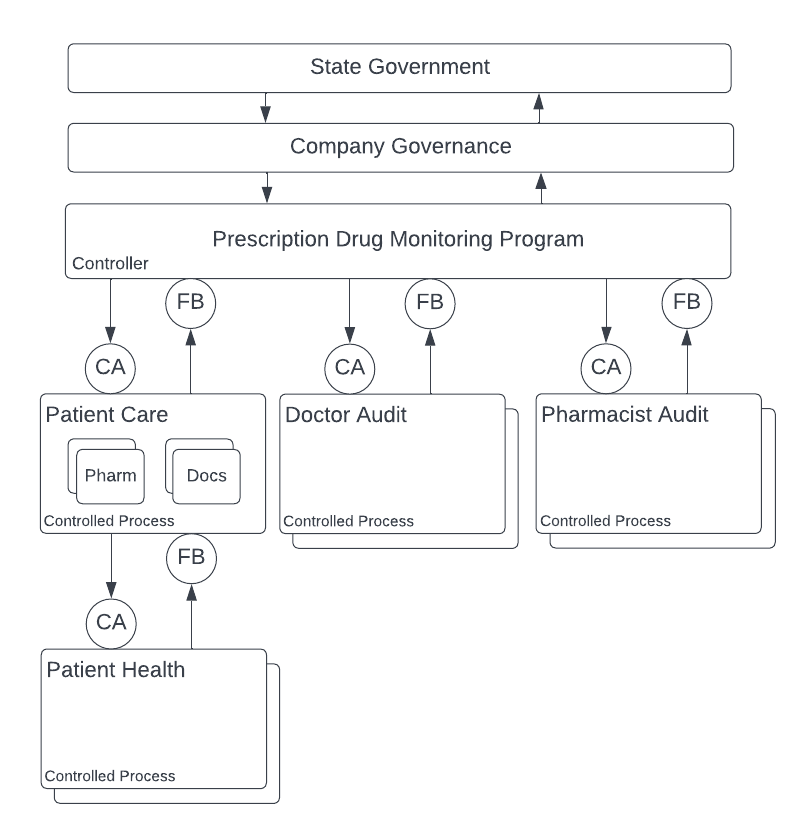} 
\caption{High Level Control Structure}
\label{PDMP_ctrl}
\end{subfigure}
\begin{subfigure}{0.5\textwidth}
\includegraphics[width=0.9\linewidth, height=6cm]{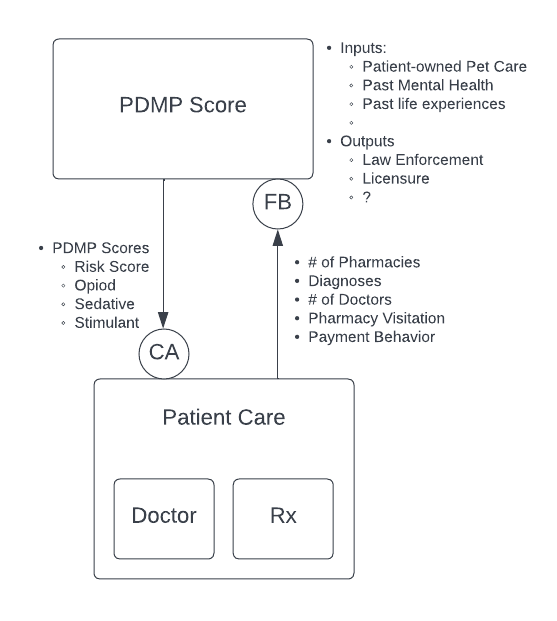}
\caption{Patient Care Control Loop}
\label{care}
\end{subfigure}

\caption{PDMP Score Control Structure and Inset Patient Care Loop}
\label{ovrall_W_inset}
\end{figure}

\begin{figure}[h!]
     %\centering
     \begin{subfigure}[b]{0.5\textwidth}
         %\centering
         \includegraphics[width=\textwidth]{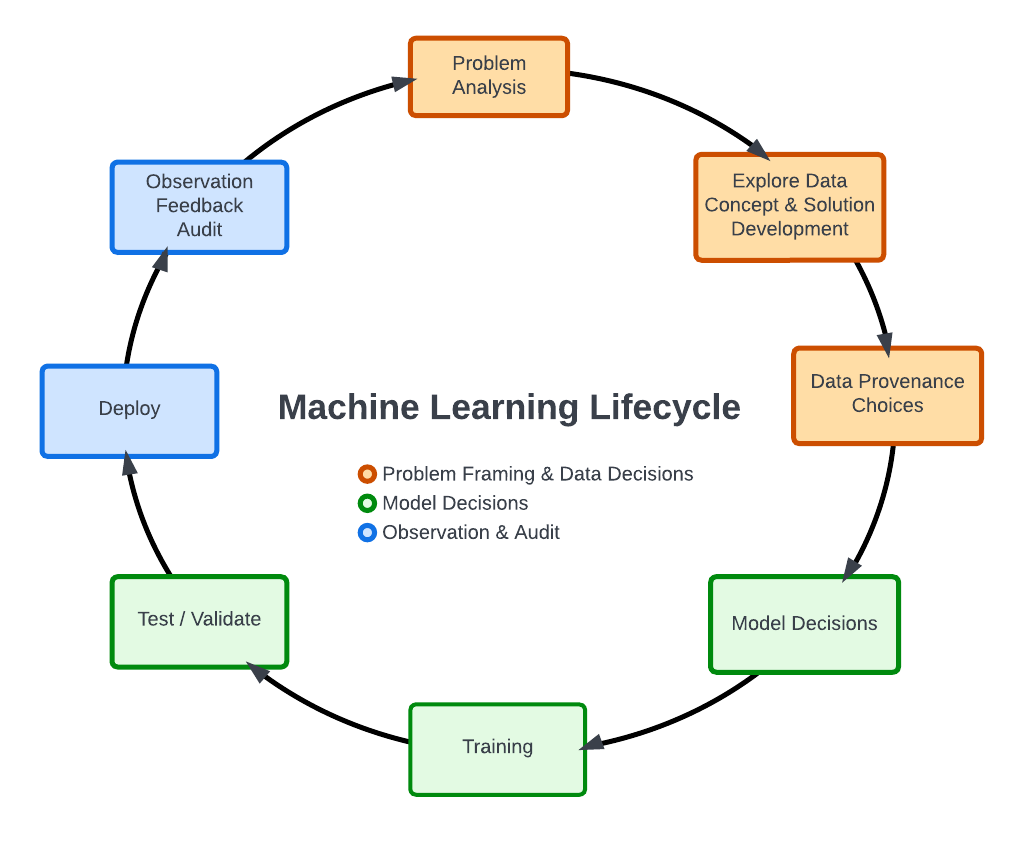}
         \caption{ML Life-Cycle}
         \label{cycle}
     \end{subfigure}
     \hfill
     \begin{subfigure}[b]{0.5\textwidth}
         %\centering
         \includegraphics[width=\textwidth]{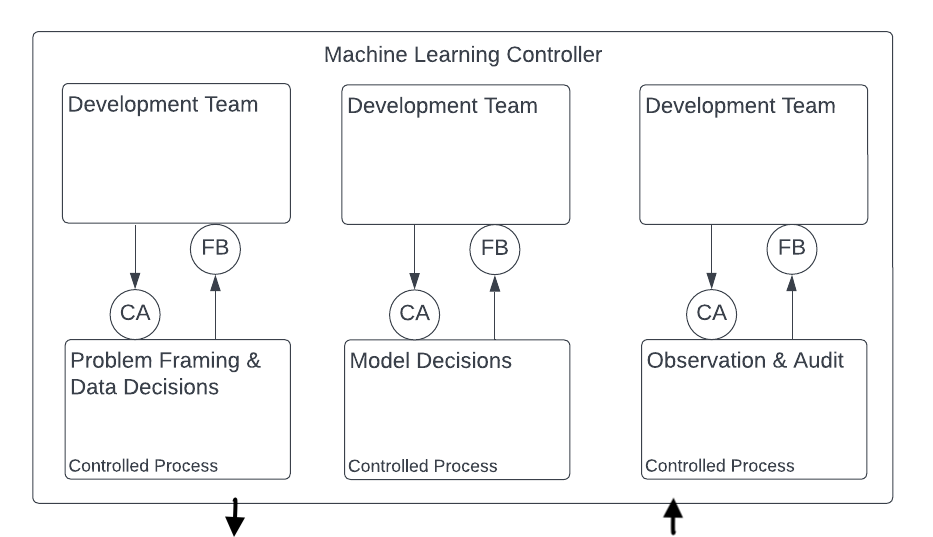}
         \caption{PDMP Cycle: Control Structure}
         \label{dev}
     \end{subfigure}
     \caption{Modeling Life-Cycle Control}
     \label{cycle-treatment}
\end{figure}

\begin{figure}[h!]
    \centering
    \includegraphics[width=0.5\textwidth]{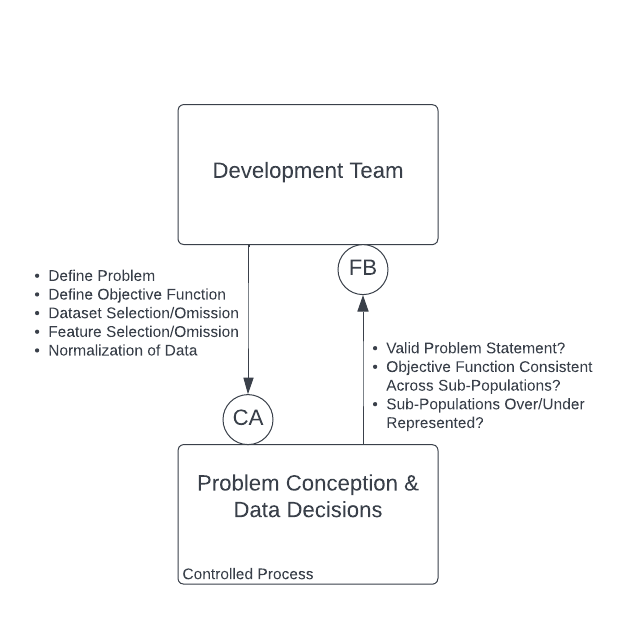}
    \caption{Data Decisions}
    \label{data}
\end{figure}

Figure \ref{PDMP_ctrl} situates our identified operational control loops for the PDMP scoring algorithm within the health system. Figure \ref{care} shows more detail for the particular Patient Care control loop for the PDMP score. Figures \ref{cycle}, \ref{cycle-treatment} and \ref{data} show the machine learning lifecycle, how this study proposes to model that cycle as a set of control loops and the specific control loop addressing the Problem Conception and Data Decisions portions of the cycle, respectively.  

     \hfill

\subsection{Control Loop Component Descriptions} \label{cont_det}
\textbf{PDMP Score Algorithm}: Feedback and Control Actions
\textit{Patient} - Given patient data, provide risk score for propensity for opioid abuse, overdose, diversion to inform healthcare clinical decisions regarding treatment and prescriptions. 
\begin{itemize}
    \item \textit{PDMP CA}: Individual risk score and component scores for narcotics, stimulants and benzodiazepines.
    \item \textit{Data Feedback}: prescriptions, diagnoses, medical history, doctor selection, pharmacy selection, arrest history, prescription payment behavior, pet prescriptions, ?
\end{itemize}
\textit{Doctor} - Track milligram morphine equivalents (MME)s prescribed per patient and other stats in order to flag 'aberrant' behavior associated with risk for enabling or participating in drug diversion--deliver letters warning practitioners when thresholds crossed and later flag to law enforcement if and when mandated by state law.
\begin{itemize}
	    \item \textit{PDMP CA}: Average PDMP risk score of Patients, average MME prescribed, stats as defined by state and DEA tracking.
	    \item \textit{Doctor Feedback}: prescriptions, retained patients, prescribing amounts compared to standards (not tailored to patient health)
\end{itemize}
\textit{Pharmacist} - require consideration of score before filling prescription, consult prescribing doctor, reserve right to refuse patient prescription if deemed too risky.
\begin{itemize}
	    \item \textit{PDMP CA}: avg PDMP risk score?. Average MMEs/ prescription? How are Pharmacists audited for their part in the system?
	    \item \textit{Pharm Feedback}: Prescriptions filled, workflow executed, reports. 
\end{itemize}

\textbf{Development Team}: This team is responsible for executing the initial phases of the ML Lifecycle.  Thus it defines the problem whose solution is sought, the conceptualization and initial decision making in forming a solution, from Operationalization of the unobservable theoretical construct to the data provenance decisions and model decisions. Requirements set. Why system is needed. Problem to solve.

\section{Team Composition and Study Limitations} \label{team}
Our analysis team is composed of five ML researchers, one sociologist, and one STPA subject matter expert. In addition to working with an experienced STPA expert, each member of the team worked to become thoroughly familiar with STPA by reading core STPA expositions -- Leveson's \textit{Engineering a Safer World} textbook~\cite{leveson_engineering_2016}, an associated \textit{STPA Handbook}~\cite{leveson_stpa_handbook_2018}, and training videos from the group that developed the framework~\cite{thomas_2021} -- and studying the STPA evaluation literature.  %It turns out that despite this effort, the meta aspect of our analysis \textit{of STPA} while doing an STPA was harder than anticipated, requiring us to revisit process documents throughout, as we sought to faithfully execute STPA where possible and yet explore options when some aspect did not neatly fit. As noted above, the STPA was conducted on a PDMP scoring algorithm. 
We did not have direct access to PDMP scoring systems or their datasets, however we were able to derive necessary details from from state-issued systems operating manuals, pharmacy work-flows, legal requirements, training documents as well as recent research involving the re-creation and testing of ML models based on the datasets and features stated by proprietors to be used and most "predictive of unintentional overdose death"~\cite{Kilby_algorithm, indiana_narx_manual, texas_narx_manual,NC_narx}. Given the top-down framing of the STPA approach, these documents proved sufficient to provide ample operating and development context to model control structures and for an effective exploration of STPA analysis for social and ethical impact.

\section{Losses and Hazards}
In our initial analysis we identified over thirty potential social, ethical and safety losses that could result both from the PDMP score system proper or downstream and yet a result of the PDMP score system's interactions within the broader healthcare system, these are fully outlined in Appendix \ref{initial_loss}. To simplify analysis going forward it was then necessary to reduce this loss list to five general categories as defined in Appendix \ref{distilled}.  
\subsection{Initial Losses}\label{initial_loss}
\begin{enumerate}
    \item Patient Death
    \item Inequity between social groups
    \begin{enumerate}
        \item Allocative - Disparity in PDMP risk score can result in a disparity in: 
        \begin{itemize}
            \item Health Treatment: affecting subsequent opportunity as well if resulting treatment disparity is debilitating reducing ability to hold a job or care for children or adult dependents.
           \item Job Opportunity: Is PDMP risk score specifically prohibited from being considered when seeking a drug-dispensing or other related health care job? Can the score or some subset be an input to other products such as background checking systems, credit or hiring algorithms?
        \end{itemize}
        \item Representational - A grouping with an inappropriately high score may have the effect of categorizing a patient inappropriately as more likely drug-seeking.
        \item Quality of Service - More difficult interactions, extra intrusive questions, when interacting with Doctors, health staff and pharmacies. 
        \begin{enumerate}
            \item Alienation:
            \begin{itemize}
                \item Turned away at Pharmacies: Resulting in adverse emotions, distrust and exclusion from the benefits offered others for health treatment.
                \item Turned away as a new patient: same as above
                \item Dismissed as a patient: same as above
            \end{itemize}
            \item Increased Labor: Above reasons in Alienation repeated here as all result in additional labor for the patient to overcome to get appropriate treatment.
	        \item Service or Benefit Loss: for same reasons in alienation, benefit of treatment is lost when it cannot be overcome or cost/effort required is too high to fight.
        \end{enumerate}
    \end{enumerate}
    \item Patient has untreated pain - Physical, mental anguish, social damage, 
    \begin{enumerate}
        \item Mental Health
        \item Physical debilitation
        \item Social Damage
        \item Occupation Damage
        \item Family Care Damage
    \end{enumerate}
    \item Loss of Safe access to Treatment/Care (Abandonment): 
    \item Behavior Herding: Desperate, deeply affected individuals may be herded to get the care they need from illegal means, thereby increasing risk of incarceration, addiction, abuse and death as the illegal treatment has no protections from overdose or doctor and pharmacist oversight.
    \item Loss of patient care (narcotic, benzodiazepines, stimulants; overall)
    \item Degraded Quality of Life: Loss of ability to work, care for children, enjoy normal life, care for adult dependents. 
    \item Law enforcement action - See CA state review -- "law enforcement surveillance and its attendant threat of criminal investigation and prosecution incentivize patient abandonment, forced taper, and involuntary medication discontinuation.
    \item Reputation loss 
    \item Privacy Violations
    \item Licensure (Doc/Pharm)
    \item Increased Liability Insurance (Docs and Pharm)
    \begin{enumerate}
        \item Social Control
        \item Financial
    \end{enumerate}
    \item Loss of Autonomy, clinical judgment 
    \item Inequity with social groups (poor people may have higher scores given method of payment is a factor)
    \begin{enumerate}
        \item Sexual assault survivor
        \item Prior arrest history
        \item Age, socioeconomic, regional, race, gender, sexuality
    \end{enumerate}
    \item Reduced accessibility to Doctors
\end{enumerate}
%\fi

\subsection{Reduced Loss List}\label{distilled}
\begin{enumerate}
    \item Death, Injury or Disability:
    \begin{itemize}
        \item Patient Death
        \item Untreated Medical Conditions (Pain)
        \item Additional Physical or Mental Injury
    \end{itemize}
    \item Disparity of Benefit/Harm
    \begin{itemize}
        \item Allocative Disparity
        \item Representational Disparity
        \item Quality of Service Disparity
    \end{itemize}
    \item Social/Economic Injury
    \begin{itemize}
        \item Damage to Reputation
        \item Occupational Damage
        \item Family Damage 
        \item Privacy Violations
    \end{itemize}
    \item Damage to Quality of Healthcare
    \begin{itemize}
        \item Abandonment
        \item Loss of Autonomy in Clinical Judgement
        \item Loss of Opportunity for Care, i.e., reduced accessibility to Doctors
    \end{itemize}
    \item Coerced Criminality or Unsafe Treatment
    \begin{itemize}
        \item Herding to Unsafe/Illicit Behavior
        \item Increase in Law Enforcement Scrutiny
    \end{itemize}
\end{enumerate}

\subsection{PDMP Score Hazards}\label{hazards}
This table lists all of the identified PDMP Scoring system hazards cross-referenced with the potential losses which may result from those hazards.
\begin{table}[h!]
\begin{tabular}{ |p{0.25cm}|p{6cm}||p{0.5cm}|p{0.5cm}|p{0.5cm}|p{0.5cm}|p{0.5cm}|  }
 \hline
 \multicolumn{7}{|c|}{PDMP Risk Score Hazards} \\
 \hline
 &Hazards&L1&L2&L3&L4&L5\\
 \hline
 1&Over-prescribe   &X&&&X&\\
 \hline
 2&Under-prescribe &X&&X&X&\\
 \hline
 3A&Inappropriately Scored - High &X & &X &X&X\\
 \hline
 3B&Inappropriately Scored - Low &X & &  &X&\\
 \hline
 4&Score Leaked &X&X&X&X&X\\
 \hline
 5&Problematic/Biased Data & &X&&X&\\
 \hline
 6&Abandonment &X&&X&X&X\\
 \hline
 7&Not provided most effective treatment.&X&&X&X&X \\
 \hline
 8&Patient gives up on medical system.&X&&X&&X \\
 \hline
 9&Excessive false positives.&X&X&X&X& \\
 \hline
 10&Excessive false negatives.&X&&&X& \\
 \hline
\end{tabular}
\caption{PDMP Risk Score System Hazards}\label{haz_tbl}
\end{table}

\section{PDMP Score Unsafe Control Actions}\label{UCA}
\begin{table}[h!]
\begin{tabular}{ |p{1.25cm}||p{3cm}|p{2.5cm}|p{0.5cm}|p{3cm}|p{3cm}|  }
 \hline
 \multicolumn{6}{|c|}{UCAs: Patient Care Control Loop} \\
 \hline
 Control Action&Not Provided&Provided&TE TL&Too Low&Too High\\
 \hline
Risk Score & Score defaults to zero. Hazard if patient susceptible to addiction - H1&H6, H7, H8&N/A&H1, H3, H10&H2, H3A, H6, H7, H8, H9\\

 \hline
\end{tabular}
\caption{Patient Care: Unsafe Control Actions}
\label{UCA_care}
\end{table}

\begin{table}[]
\begin{tabular}{ |p{3cm}||p{0.5cm}|p{2.5cm}|p{0.5cm}|p{3cm}|p{3cm}|  }
 \hline
 \multicolumn{6}{|c|}{UCAs: Problem Conception and Data Decision Control Loop} \\
 \hline
 Control Action&Not Prov&Provided&TE TL&Too Few/Little&Too Many/Much\\
 \hline
Define Problem &N/A&H5, H3&N/A&N/A&N/A\\
Define Obj. Function&N/A&H1-3, H5-10&N/A&N/A&N/A\\
Dataset Selection or Omission&N/A&H1-3, H5-10&N/A&N/A&N/A\\
Feature Selection or Omission&N/A&H1-3, H5-10 &N/A&H1-3,H6-10 &H1-3,H6-10\\
Data Normalization&N/A&H3, H5, H9, H10&N/A&H3, H5, H9, H10&H3, H5, H9, H10\\
 \hline
\end{tabular}
\caption{Problem Conception and Data Decisions}
\label{UCA_dev}
\end{table}

\section{Subject ML System: Prescription Drug Monitoring Program (PDMP) Score} \label{PDMP}
Prescription Drug Monitoring Programs are mandated in all 50 states and are intended to prevent or curtail widespread healthcare issues such as drug addiction, misuse and overdose deaths~\cite{narxcare_webpage}. A majority of these programs employ a risk scoring system as a clinical tool, requiring physicians, pharmacists and their staffs to review a patient's risk scores prior to writing or filling prescriptions for certain schedules of drugs~\cite{west_virginia_2018, indiana_narx_manual}. These risk scores are calculated by machine learning systems trained on a variety of data sources~\cite{texas_narx_manual,kilby_economics_2016}. Thus, these ML-based PDMP tools assist in governing the health care of hundreds of millions of people across the United States. In systems like this one, which affect large numbers of people in highly consequential ways, impacting life, health and livelihood, it is incumbent on developers, company management and government officials to demonstrate due diligence by showing evidence that an ML-enabled system not only improves the performance of the sociotechnical system which it is augmenting, but that it does not also introduce unacceptable negative social and ethical impacts down stream of the system~\cite{bender_dangers_2021, kroll_accountability_2020, raji_ai_2021, martin_participatory_2020}. Often, this type of analysis is not done, is attempted ad-hoc, or is treated as if it were impossible due to its complexity~\cite{martin_extending_2020}. %In pursuit of making such an assessment of ML-based PDMP scoring systems verifiably possible, we apply System Theoretic Process Analysis to identify where potential social and ethical impact losses do arise and thus enable requirements to be levied within the scope of the system to eliminate or sufficiently mitigate those losses. In the event they cannot be sufficiently controlled from within the scope of the system considered, then use STPA to suggest external controls, provide analysis to help inform decisions regarding whether the added risk of loss is justified given the system's benefits.

\section{ML Harms} \label{harms}
A steadily growing number of incidents and calls to action demonstrate the necessity to include social and ethical impact analysis as a key component of the ML system development life cycle~\cite{bender_dangers_2021,politico_dutch_2022,perkowitz_bias_2021,Angwin_arrest_recidivism_2016}. However, in order to enable such an assessment of social and ethical impacts, we must first begin with a firm understanding of the various harms that can result from sociotechnical algorithmic systems. Additionally, special attention must be given to the deployment environment as sociotechnical systems often have far reaching social and technological connections and impacts for humans which can result in losses, hazards, and negative outcomes for entities far removed from the system--in the case of PDMP scoring systems these will no doubt include patients, doctors, and pharmacists, but also patient families and the functionality of and trust in the health system at large. 

This paper adopts Shelby \textit{et al}.'s recent work which successfully taxonomized the myriad manifestations of algorithmic harms~\cite{shelby2023harms}. This taxonomy provides an initial yet robust foundation to proceed from, and we use it as a basis for developing our subject system's social and ethical losses and hazards, a key part of the first step of STPA.

\section{STPA for Social and Ethical Impact} \label{SEI}
Although researchers suggest existing safety frameworks can address concerns of social and ethical impact in ML~\cite{dobbe_system_2022, li_fmea-ai_2022}, it is not known whether they are effective. Moreover, studying social and ethical risks is particularly difficult because they are often substantially decoupled from the individual components which are typically the objects of analysis~\cite{selbst_fairness_2019, martin_extending_2020}. Thus, this research investigates whether such frameworks, STPA in particular, cause the identification of the social and ethical risks to surface as a natural consequence of their process, and further if they likewise necessarily offer a valid path (assuming one exists) to sufficiently correct and control for the discovered hazardous states.  

This research will perform STPA on a machine learning sociotechnical system to determine STPA's effectiveness in social and ethical impact analysis and correction.  It accomplishes this by treating social and ethical harms as losses to determine if STPA recovers a useful set of hazard scenarios.  Moreover, it is the focus of the case study to determine how effective STPA is at identifying corrective actions to prevent (or mitigate) the discovered negative social and ethical impacts of the ML system. As we continue our research we will take these steps and apply them to additional subject systems from other ML system domains, such as those leveraging large language models or machine vision classification systems.  In so doing we hope to introduce a tool to provide an organized, proven systematic approach to improved social and ethical analysis for complex ML-enabled sociotechnical systems.

\end{document}